\documentclass[11pt,a4paper]{article}
\usepackage[hyperref]{eacl2021}
\usepackage{times}
\usepackage{latexsym}
\usepackage{times}
\usepackage{epsfig}
\usepackage{amsmath}
\usepackage{amssymb}
\usepackage{latexsym}
\usepackage{graphicx}
\usepackage{multirow}
\usepackage{breqn}

\usepackage{graphicx}
\usepackage{amsmath,graphicx}
\usepackage{times}
\usepackage{latexsym}
\usepackage{times}
\usepackage{latexsym}
\usepackage{caption}
\usepackage{subcaption}
\usepackage{comment}
\usepackage{times}
\usepackage{soul}
\usepackage{booktabs}
\usepackage{url}
\usepackage[utf8]{inputenc}
\usepackage{graphicx}
\usepackage{amsmath}
\usepackage{amsmath,graphicx}
\usepackage{times}
\usepackage{latexsym}
\usepackage{times}
\usepackage{latexsym}
\usepackage{comment}
\usepackage{times}
\usepackage{soul}
\usepackage{booktabs}
\usepackage{url}
\usepackage[utf8]{inputenc}
\usepackage{graphicx}
\usepackage{amsmath}
\usepackage{amssymb}
\usepackage{array}
\usepackage{booktabs}
\usepackage{multirow}
\usepackage{multicol}
\usepackage{caption}
\usepackage{xspace}
\usepackage{amssymb}
\usepackage{booktabs}

\usepackage{multirow}
\usepackage{multicol}
\usepackage{caption}
\usepackage{xspace}
\usepackage{amsmath}

\usepackage{tabularx}
\usepackage{algorithm2e}

\DeclareSymbolFont{rsfs}{U}{rsfs}{m}{n}
\DeclareSymbolFontAlphabet{\mathscrsfs}{rsfs}
\long\def\symbolfootnote[#1]#2{\begingroup%
\def\thefootnote{\fnsymbol{footnote}}\footnote[#1]{#2}\endgroup}

\usepackage{microtype}

\aclfinalcopy 
\def\aclpaperid{31} 

\title{Analyzing Curriculum Learning for Sentiment Analysis along Task Difficulty, Pacing and Visualization Axes}

\author{Anvesh Rao Vijjini* \\
\\\And
Kaveri Anuranjana* \\ \\
 International Institute of Information Technology - Hyderabad \\
 \texttt{\{vijjinianvesh.rao,kaveri.anuranjana\}@research.iiit.ac.in} \\ \texttt{radhika.mamidi@iiit.ac.in} \\ \And
 Radhika Mamidi\\ 
 \\}
 
\date{}
\begin{document}
\maketitle
\begin{abstract}
While Curriculum Learning (CL) has recently gained traction in Natural language Processing Tasks, it is still not adequately analyzed. Previous works only show their effectiveness but fail short to explain and interpret the internal workings fully. In this paper, we analyze curriculum learning in sentiment analysis along multiple axes. Some of these axes have been proposed by earlier works that need more in-depth study. Such analysis requires understanding where curriculum learning works and where it does not. Our axes of analysis include Task difficulty on CL, comparing CL pacing techniques, and qualitative analysis by visualizing the movement of attention scores in the model as curriculum phases progress. We find that curriculum learning works best for difficult tasks and may even lead to a decrement in performance for tasks with higher performance without curriculum learning. We see that One-Pass curriculum strategies suffer from catastrophic forgetting and attention movement visualization within curriculum pacing. This shows that curriculum learning breaks down the challenging main task into easier sub-tasks solved sequentially.

\end{abstract}

\section{Introduction}
\let\thefootnote\relax\footnotetext{*The authors contributed equally to the work.}
Learning in humans has always been a systematic approach of handling the fundamentals first and then learning incrementally harder concepts. Cognitive Science has established that this leads to a clearer, robust understanding and the most efficient learning 
\cite{krueger2009flexible,avrahami1997teaching}. Indeed, something similar can be applied while training neural networks. \cite{bengio2009curriculum} show that Curriculum Learning (CL) - sampling data based on increasing order of difficulty leads to quicker generalization.
\cite{weinshall2018curriculum} also demonstrate that CL increases the rate of convergence at the beginning of training. Their CL strategy involved sorting the training data based on transfer learning from another network trained on a larger dataset. The idea of reordering samples has been explored in various approaches.
In this paper, we evaluate the ``easiness'' with a network and train the samples on another network. Hence, even in our case, we shall pick easier points regarding a target hypothesis then train another network that optimizes its current hypothesis. This idea has been suggested by previous works as well. \cite{hacohen2019power,weinshall2018curriculum} 

 \cite{cirik2016visualizing} proposed Baby Steps and One Pass curriculum techniques using sentence length as a curriculum strategy for training LSTM \cite{hochreiter1997long} on Sentiment Analysis. A tree-structured curriculum ordering based on semantic similarity is proposed by \cite{han2017tree}. \cite{rao2020sentiwordnet} propose an auxiliary network that is first trained on the dataset and used to calculate difficulty scores for the curriculum order. 
 
 CL is also used by NLP in tasks like Question Answering \cite{sachan2016easy,sachan2018self} and NLG for Answer Generation \cite{liu2018curriculum}. For Sentiment Analysis, \cite{cirik2016visualizing} propose a strategy derived from sentence length, where smaller sentences are considered easier and are provided first. \cite{han2017tree} provide a tree-structured curriculum based on semantic similarity between new samples and samples already trained. \cite{bayesian2016learning} suggest a curriculum based on handcrafted semantic, linguistic, syntactic features for word representation learning. 
 
 Some of these works \cite{cirik2016visualizing,han2017tree,rao2020sentiwordnet} have suggested that Baby Steps performs better than One Pass. We perform experiments using both techniques. While the idea of curriculum remains the same across these works, the strategy itself to decide sample ordering is often diverse.

\section{Axis I: Curriculum Learning: One Pass and Baby Steps}
\label{sec:baby_steps}
While Curriculum Learning as defined by \cite{bengio2009curriculum} is not constrained by a strict description, later related works \cite{cirik2016visualizing,han2017tree,spitkovsky2010baby,rao2020sentiwordnet} make distinctions between Baby Steps curriculum and One-Pass curriculum. Most of these previous works have also shown the dominance of Baby Steps over One-Pass. Baby Steps and One Pass curriculum can be defined as follows. 
For every sentence $s_{i}$ $\in$ $D$, its sentiment is described as $y_{i}$ $\in$ $\{0,1,2,3,4\}$\footnote{Our dataset has 5 labels.}, where $i$ $\in$ $\{1,2...n\}$ for $n$ data points in $D$. For a model $f_{w}$, its prediction based on $s_{i}$ will be $f_{w}(s_{i})$. Loss $L$ is defined on the model prediction and actual output as $L(y_{i},f_{w}(s_{i}))$ and Cost defining the task as $C(D,f_{w})$ as 

\begin{equation}
\sum_{\forall i} \frac{1}{n}L(y_{i},f_{w}(s_{i}))    
\end{equation}
 
 Here, curriculum strategy $S(s_{i})$ defines an ``easiness''/``difficulty''  quotient of sample $s_{i}$. Furthermore, One Pass makes distinct, mutually exclusive sets of the training data and trains on each one of these sets one by one. This makes it faster as compared to Baby Steps, where data cumulatively increases in each pass. This implies that the model is trained on previous data and the additional harder data.

\begin{figure}
\centering
  \includegraphics[width=7.5cm]{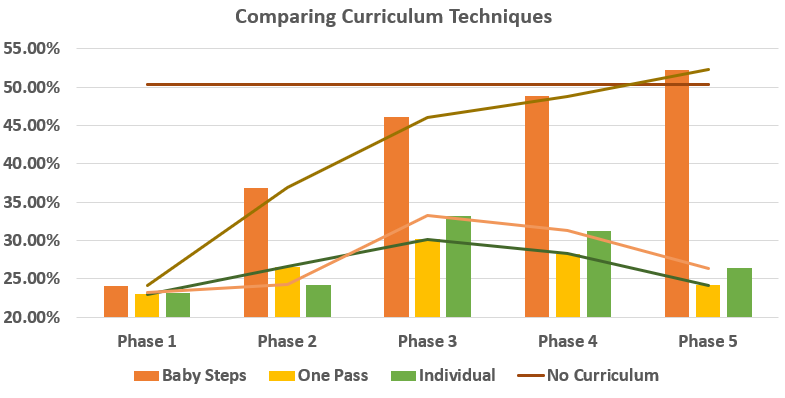}
  \caption{Performance comparison shows weakness of One Pass.}
  \label{fig:comparison}
\end{figure}

To analyze the two methods for executing CL we choose two curriculum strategies (difficulty scoring function). Furthermore, we also experiment with an individual setting explained in following sections.
\subsection{Dataset}
Following previous works in curriculum-driven sentiment analysis \cite{cirik2016visualizing,han2017tree,bayesian2016learning,rao2020sentiwordnet} We use the Stanford Sentiment Treebank (SST) dataset \cite{socher2013recursive}\footnote{https://nlp.stanford.edu/sentiment/}. Unlike most sentiment analysis datasets with binary labels, SST is for a 5-class classification consisting of 8544/1101/2210 samples in train, development, and test set. We use this standard split with reported results averaged over 5 turns. 
\subsection{Model Details: BERT}
We use the popular transformer model BERT\cite{devlin2018bert} for our experiments due to how ubiquitous it is across natural language processing tasks. Bidirectional Encoder Representations from Transformers (BERT) is a masked language model trained on large corpora. A sentence is added with a special token (CLS) at the beginning and is passed into the pre-trained BERT model. It tokenizes the sentence with a maximum length of 512 and outputs a contextual representation for each of the tokenized words. There are variants of pre-trained BERT depending upon the hyper-parameters of the model. BERT-Base Uncased consists of 12 transformer encoders, and output from each token is a 768 dimension embedding. We apply a softmax layer over the first token position (CLS) output by the BERT model for sentiment analysis.

\subsection{Curriculum Strategies}
\subsubsection{Auxiliary Model Strategy}
\label{sec:aux_strategy}
Auxiliary Model Strategy is based on previous works \cite{weinshall2018curriculum,hacohen2019power,rao2020sentiwordnet} which propose a difficulty scoring function, transfer learned from an another network. We first train an auxiliary model $Aux$ for sentiment analysis on the same dataset. This $Aux$ model architecture will be the same as the model finally used for CL. This allows us to find out which training samples are actually difficult. 
We learn what samples are the most difficult to classify and what are the easiest from this model. For all training samples of $D$, we define the curriculum score as follows:
\begin{equation}
S(s_i) = \sum_j^c(Aux(s_i)^j - y_i^j)^2
\end{equation}
where $Aux(s_i)^j$ is the prediction of auxiliary model $Aux$ on sentence $s_i$, $j$ is the iterator over the number of classes $c=5$. In essence, we find the mean squared error between the prediction and the sentence's true labels. If $S(s_i)$ is high, it implies the sentence is hard to classify, and if less, then the sentence is easy. Because the features were trained on an auxiliary model from BERT features, we get an easiness-difficulty score purely from the perspective of sentiment analysis. Section \ref{sec:examples} gives some examples of difficult and easy samples according to this curriculum. 
\subsubsection{Sentence Length}
This simple strategy tells that architectures, especially LSTM find it difficult to classify longer sentences. Hence, longer sentences are difficult and should be ordered later. Conversely, shorter sentence lengths are easier and should be trained first. This strategy is prevalent and has not only been used in sentiment analysis \cite{cirik2016visualizing}\footnote{\cite{cirik2016visualizing} have done on CL SST as well. However, our numbers do not match because they use the much larger phrase dataset.} but also in dependency parsing \cite{spitkovsky2010baby}. This is why it becomes a strong comparison metric, especially to evaluate the distinction between One Pass and Baby Steps.

\subsubsection{Individual}
In this strategy, we report accuracy on the test set when trained on $D_{1}$,$D_{2}$, and so on individually. This is by no means a curriculum strategy since no model ever sees the complete training data. The Individual experiment can be thought of as One Pass, but the weights are reset after every training phase. 

\begin{table*}[t]
  \centering
  \setlength{\tabcolsep}{4pt}
  \renewcommand{\arraystretch}{1.25}
  \begin{tabular}{{c c c c}}
    \hline
     Model&  & Curriculum Strategies \\
    & Baby Steps & One Pass  & No Curriculum\\
    \hline
     Auxiliary Strategy& 52.3 &\textbf{24.15} & 50.03 \\
    \hline
    Sentence Length Strategy  & 51.2 &\textbf{23.64} & 50.03\\
    \hline
     \hline
  \end{tabular}
  \caption{Accuracy scores in percentage of the two curriculums across Baby Steps and One Pass as compared to no curriculum on the SST5 dataset.}
  \label{tab:Accuracy}
\end{table*}

\subsection{Results and Discussion}
Table \ref{tab:Accuracy} and Figure \ref{fig:comparison} give us an insight into what is happening with One Pass. Firstly, Table \ref{tab:Accuracy} clearly shows that One Pass underperforms both No curriculum and Baby Steps for both the Curriculum Strategies. This is in line with previous experiments as well \cite{cirik2016visualizing,han2017tree,rao2020sentiwordnet}. One Pass is less time-consuming since it only observes a sample once, unlike Baby Steps which repeatedly sees samples from the previous and the current phase. Furthermore, we see that auxiliary outperforms sentence length. Finally, Figure \ref{fig:comparison} illustrates the reason for One Pass's weakness. We observe that on successive phases, One Pass closely follows the curve of the proposed Individual experiment. However, unlike One Pass, Individual has no memory of samples at previous stages. This shows that in every phase of One Pass, the model forgets the previous stage samples and effectively behaves like Individual, hence catastrophically forgetting previous phases. Furthermore, this is especially a major issue for the myriad of methods involving the Language Model pre-training and downstream tasks fine-tuning Paradigm. Language Model backed transformers \cite{vaswani2017attention} are now ubiquitous for tasks across Natural Language Understanding \cite{devlin2018bert,liu2019roberta, yang2019xlnet, lan2019albert,raffel2019exploring,choudhary2020self}. These architectures are trained in a language model task first, followed by fine-tuning on the downstream task. In this regard, they are similar to the One Pass strategy since they successively train on disjoint datasets. Hence, problems with One Pass, such as catastrophic forgetting, are likely to occur with these architectures as well. Additionally, while Baby Steps addresses the catastrophic forgetting in One Pass, it will be harder to address catastrophic forgetting in Language Models. The downstream task objective in the CL setting is different from the Language Model objective making joint training harder, unlike Baby Steps.

\subsubsection{Visualizing Catastrophic Forgetting}
\begin{figure*}
\centering
  \includegraphics[width=15cm]{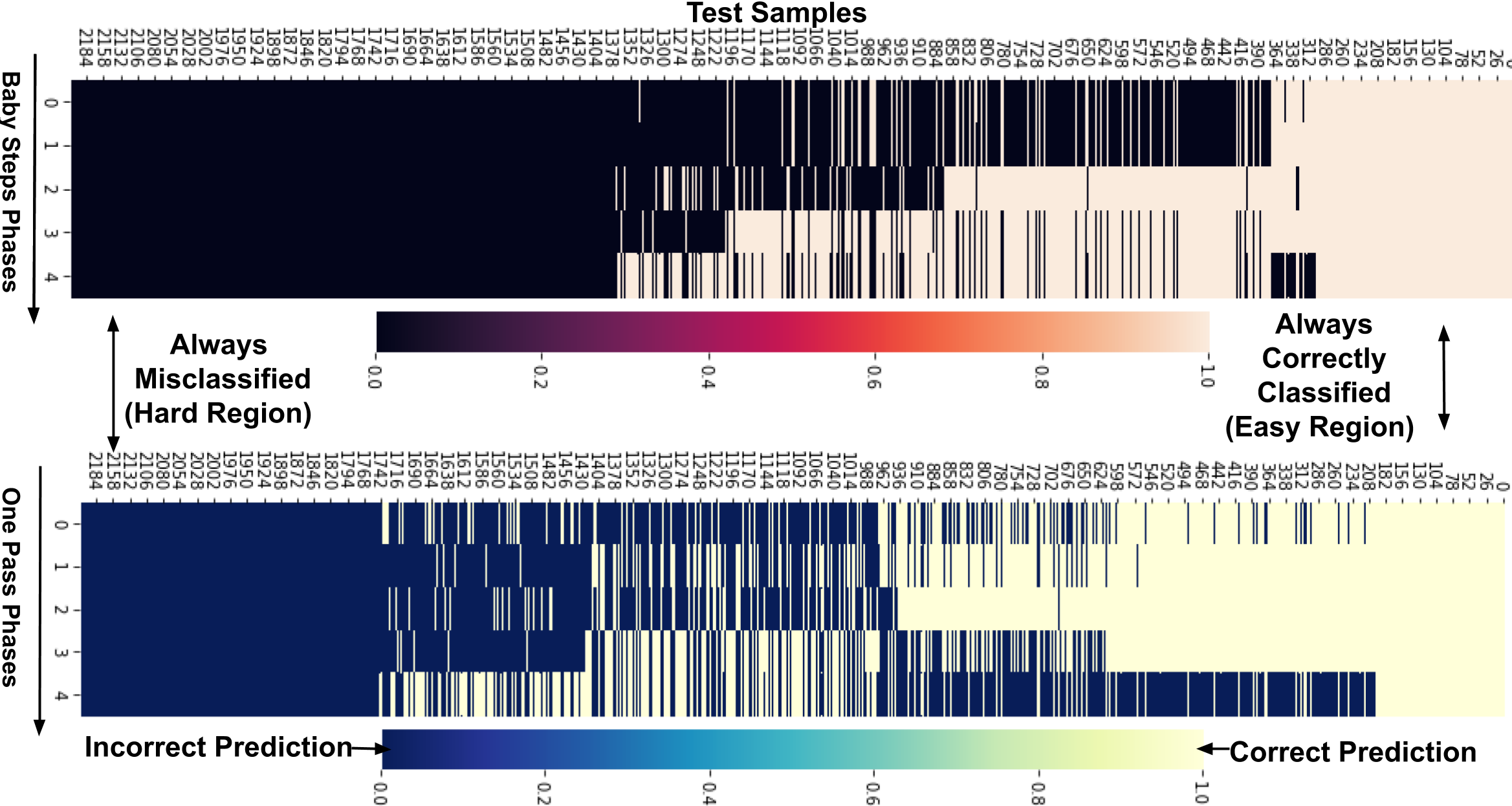}
  \caption{Catastrophic Forgetting in One Pass Visualized. In the above images, the model performance on each individual test set sample is illustrated across all curriculum phases. These phases correspond with how much data the model has observed so far.}
  \label{fig:cata_viz}
\end{figure*}

To visualize Catastrophic Forgetting, Figure \ref{fig:cata_viz} plots the BERT model's performance. The figure illustrates the corresponding match or mismatch between model prediction and ground truth rather than just the model prediction. In this figure, the correctness of model prediction across all test samples is illustrated for every phase of the One Pass and Baby Steps methods. The samples are vertically stacked along the $y$-axis in an order based on the number of phases sample missclassified in. The consecutive phases of the curriculum training are indicated on the $x$-axis.\footnote{As discussed earlier, we take $k = 5$; hence we end up with 5 phases on the $x$-axis.}. A darker color (value of ``0") indicates miss-classification, and brighter or lighter color (value of ``1") indicates that the sample is correctly classified. Note that the classification task itself is not a binary task but a multi-class classification problem. For SST5, this is a five-class classification task. Figure \ref{fig:cata_viz} edifies the following points.
\begin{itemize}
    \item \textbf{Easy and Hard Samples}: In both the figures, some sections are always dark or always light in color. The difficulty of certain samples is consistent irrespective of model training. The accuracy is actually determined by the samples with intermediate difficulty whose prediction can fluctuate as the model observes more data encounters. 
    \item \textbf{Memory in Baby Steps}: The figure for Baby Steps shows the model fairly remembering well the concepts it is trained on. Here, once the model prediction is corrected, it mostly stays corrected. Hence, implying a ``memory" in the model. 
    \item \textbf{Catastrophic Forgetting in One Pass} Contrary to Baby Steps, One Pass heavily suffers from catastrophic forgetting, which is observed every time the color of a sample changes from lighter to darker. In One Pass, in successive phases, the model makes more correct predictions in new regions. However, at the same time, samples become darker in the regions earlier it was lighter in. Model performance decreases to the lowest in the final stages because the model is the farthest from all previous learning phases at this point. 
    \item \textbf{Visualizing Test Performance}: This visualization is done on the unseen test set. Our above hypotheses are still natural to understand if visualized on the train set. However, the samples in the visualization are always unseen during the training. This implies that the model forgets or remembers training samples in the corresponding phases and the associated concepts as well.
\end{itemize}

\section{Axis II: Curriculum Learning only helps Difficult tasks}
\label{sec:5_difficult_tasks}
\cite{hacohen2019power} and \cite{xu2020curriculum} propose that CL might help tasks which are harder than easier. They observe that for tasks which have low performance, the difference in performance caused due to introducing a curriculum training is more than the tasks which have a higher performance without curriculum. We call this the Task Difficulty Hypothesis.

\cite{hacohen2019power} perform an image classification experiment where they experiment with enabling curriculum on the CIFAR dataset\cite{krizhevsky2009learning} with 100 classes(CIFAR-100) and 10 classes(CIFAR-10) as two separate experiments. They use VGG network\cite{simonyan2014very} as the common neural network for these experiments. Naturally, they report performance of CIFAR 10 in the range of 90 to 100$\%$ (hence, an \textit{easy} task) and CIFAR-10 in the range pf 60 to 70 $\%$ (hence, a \textit{hard} task). On addition of a curriculum training to these datasets, they observe that the increment in performance on CIFAR-100 is almost twice the increment in performance on CIFAR-10 while using the same network VGG-net for training. They argue that this might be because in easier tasks such as CIFAR-10, there are already enough easy samples observed during training without CL, and hence improvement caused by CL is subdued.

\cite{xu2020curriculum} enable their CL across the range of tasks in GLUE\cite{wang2018glue}. GLUE encompasses a wide range of tasks in natural language understanding, with varying performances. For example a task such as RTE or CoLA is considered harder than SST-2 or QNLI. The BERT\cite{devlin2018bert} model's performance for the same are 70.1, 60.5, 94.9,  91.1 respectively \footnote{As reported in the original paper \cite{devlin2018bert}}.

We built our experiments upon these works. Proposed work is different from \cite{hacohen2019power} in the way that our work focuses on sentiment analysis in NLP, whereas \cite{hacohen2019power} experimented strictly with image processing tasks. Furthermore, since the apparent relation between task difficulty and improvement due to CL wasn't the main focus of the work, the experimentation wasn't enough to fully establish the correlation. While \cite{xu2020curriculum} perform experiments within the purview of NLU and Text Classification, the experiments themselves are not consistent across dataset and the nature of task. Specifically, we believe it's hard to conclude from experiments on tasks as disparate as Linguistic Acceptability\cite{warstadt2019cola} and Sentiment Analysis\cite{socher2013recursive} to establish the relationship between CL and task difficulty. Such an experiment inadvertently raises more questions whether the difference of improvement due to CL is due to the nature of tasks (Sentiment Identification as opposed to Linguistical soundness detection) or the nature of the dataset being different (sentence lengths or vocabulary). We eschew this concern by performing experiment within the same task (Sentiment Analysis), and the same dataset (SST-5), instead we introduce change in task difficulty by using the fine grained labels provided by this dataset. This experimental setup will be explained in more detailed in the following sections.
\subsection{Dataset}

To ensure no effect of nature of task or dataset on our experiments, we utilize just a single dataset but under different conditions and sampling to simulate difficulty. All these meta-datas are generated using the SST-5 \cite{socher2013recursive}\footnote{https://nlp.stanford.edu/sentiment/} dataset. It is important to note that we do not use the phrase data for training which is why our scores may fall short of earlier reported performances of BERT on SST5 and SST derived datasets \cite{devlin2018bert,liu2019roberta,xu2020curriculum}. The fine-grained 5 classes of this dataset enable us to create easier variants of the same dataset without any significant change in the problem statement definition and the dataset statistics. We generate four datasets from SST5: SST-2, SST-3, SST-4 and SST-5 itself. In SST-3 and SST-5 neutral label is preserved, otherwise dropped. The two negative labels and the two positive labels are clubbed in SST-3 and SST-2.

In each of the SST-$x$ dataset, the train, test and dev sets are all converted accordingly. The model in all cases is strictly trained on the train data with development as validation and results are reported on test set. This way the comparison can be fair since dataset specific effects on the hypothesis would not be pertinent. 

Humans are likely to observe sentiment polarity of a natural language sentence on a continuous scale rather than distinct classes. Hence, it makes natural sense that making a distinction between, Very Positive and Positive is a harder task than just making a distinction between positive and negative.

\subsection{Curriculum Training}

We use the Auxiliary Model Strategy (Section \ref{sec:aux_strategy}) coupled with Baby Steps (Section \ref{sec:baby_steps}) as our curriculum training technique and BERT as the model architecture. We use Baby Steps \cite{cirik2016visualizing} as the curriculum pacing since we observed it to have best performance.

\subsection{Results and Discussions}
\begin{table}[t]
  \centering
  \small
  \setlength{\tabcolsep}{4pt}
  \renewcommand{\arraystretch}{1.25}
  \begin{tabular}{{c c c c}}
    \hline
    & With Curriculum & No Curriculum & Difference\\
    \hline
     SST-2 & 89.95  &  91.11 & $-$ 1.6 \\
    \hline
    SST-3 & 75.47 & 75.52 &  $-$ 0.05 \\
    \hline
    SST-4 & 62.04 & 61.72 &  $+$ 0.32 \\
     \hline
    SST-5 & 52.3 & 50.03 &  $+$ 2.0 \\
     \hline
     \hline
  \end{tabular}
  \caption{Accuracy scores in percentage on all the created meta-data with and without curriculum. Results are obtained by averaging over 5 runs.}
  \label{tab:accuracy_sst2345}
\end{table}

Our Experiments are illustrated in Figure and Table \ref{tab:accuracy_sst2345}. Following are the major points to be noted

\begin{itemize} 
\item \textbf{Number of classes is detrimental to model performance} We observe that in both with curriculum and No curriculum settings, model performance varies tremendously between SST-5 and SST-2. This is also evident in previous works which have shown performance ranges on SST to lie in the ranges of 45-50$\%$ \cite{cirik2016visualizing} for a 5 class problem and in the order of 90$\%  $\cite{lan2019albert,devlin2018bert,xu2020curriculum} for a binary class problem. Furthermore, it is interesting to note that the performance gap between SST-$i$ and SST-$i+1$ decreases as number of classes increase. This implies while adding classes makes the task harder, the hardness is less relevant if already many classes.

\begin{figure}
\centering
  \includegraphics[width=6cm]{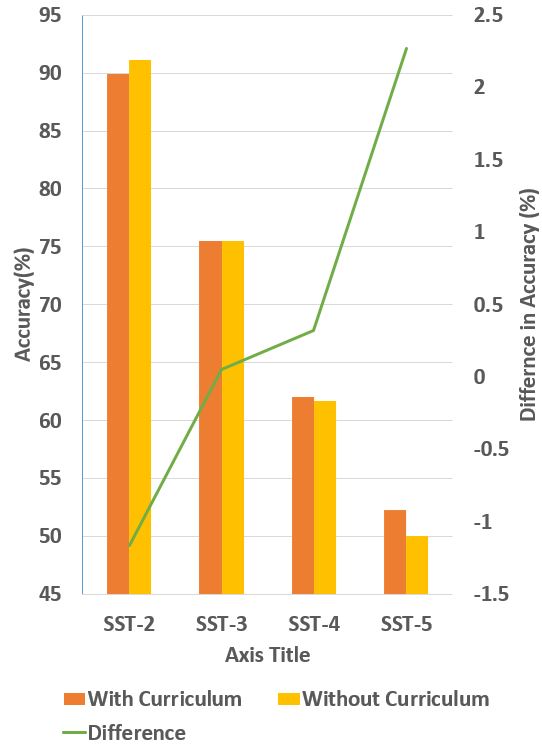}
  \caption{Accuracy scores in percentage on all the created meta-data with and without curriculum. The difference has been plotted on the secondary y-axis (Right)}
  \label{fig:meta}
\end{figure}
\item \textbf{Curriculum has adverse effects on high performing models}  The most striking point to be observed is that curriculum actually does not help in certain cases. In SST-2 and SST-3, the model is already performing quite well with accuracy of 91.11 and 75.52 respectively In these cases the model observes a decrements= of 1.6 and 0.05 $\%$ respectively due to CL. The performance on SST-2 is in line with previous work \cite{xu2020curriculum} who also observe a slight decrement to no increment in this dataset for their own curriculum technique. Tasks which are having performance are in essence, possessing higher number of easier samples than tougher tasks. In such a condition, A model observing these easy samples again and again (as we saw earlier in Baby Steps), might lead to overfitting. We believe this to be the reason behind then apparent performance degradation.
\item \textbf{Curriculum has positive effects on low performing models} Despite low performance on SST-2, the resourcefulness of CL is felt in SST-5 and SST-4 where there is a positive difference. These tasks are harder in nature with No Curriculum performance below 65$\%$, and CL is able to improve the scores significantly with an average of $+1.16\%$. As we saw in previous section, Hard samples are catastrophically forgotten all the time, hence training on these samples again and again will lead to an improvement. Unlike SST-2 and SST-3 which have a huge proportion of easy samples, Hard tasks according have harder samples and hence instead repeated training in a systematic ordering would not have an adverse effect. 
\end{itemize}

Furthermore, \cite{xu2020curriculum} suggest a similar reasoning for the Task Difficulty hypothesis. They suggest that when learning without curriculum, in the case of harder tasks, the model is confounded and overwhelmed by the presence of hard samples. Hence, laying out the training where the model observes easier samples first is natural to improve performance. However, this reasoning does not explain well why there is an apparent decrement of performance for CL on high performing tasks.

\section{Axis III: Attention Movement Visualization}
Visualization of Attention \cite{vaswani2017attention} is an important and popular visualization method for interpreting the prediction in these models. Previous works such as \cite{clark2019does}  have visualized attentions to identify where does BERT look. \cite{vaswani2017attention} also show in attention visualization that various heads of the transformer look at various linguistic details. In this section we use attention visualization to qualitatively analyze and explain how the model's focus on the sentence changes as the various stages (when model encounters new and harder data) of CL progresses. 

It is important to note that BERT has $N$ layers and $H$ heads. Within each of these $N$x$H$ heads there lies a $T$x$T$ self attention from each input time stamp to every other time stamp. Where $T$ is the maximum sentence length. Thereby, there are a total of $N$x$H$ 2D $T$x$T$ Attention visualizations observed in BERT.
\subsection{Experiments}
To particularly identify how focus of the BERT model changes across multiple stages or phases of the curriculum, we devise an attention movement visualization. We define Attention Movement index as
\begin{dmath}
M(i,h,n) = A(i,h,n) - A(i-1,h,n) \quad \forall	i \in \{1,2,3..c\}
\end{dmath}
where $M$ is the proposed Movement index, $A(i,h,n)$ is the Attention visualization of the $h^{th}$ head of $n_{th}$ layer after training $i^{th}$ phase of curriculum training and $c$ is the total number of phases in the curriculum training. We can see that a positive $M(i)$ indicates that across subsequent stages of the curriculum, the model has added attention or increased focus in the area. 

Conversely, a negative $M(i)$ indicates that after completing phase $i$ model has decided to attend to certain area or region lesser than before the phase's training. Essentially the Movement indicates relative change in attention rather than attention itself. This is significant because, unlike typical visualization where we observe a singular model in isolation, here we are attempting to analyze how the model behaves in subsequent stages of the training in a curriculum fashion and furthermore, how this change effects the prediction label.

Additionally, a point to note is that while previous works such as \cite{clark2019does} establish that BERT's distinct heads attend on linguistic notions of syntax among other linguistic ideas, we do not visualize these attention heads themselves. In this experiment, we are looking for an overall perspective into where there is a positive or negative change in the attention focus rather than understanding the individual linguistic nuances of individual heads. For this reason we further propose an Averaged Movement index as follows:
\begin{equation}
    M_{avg}(i) = \sum_{h=2,n=2}^{H,N} M(i,h,n) 
\end{equation}
where, $M_{avg}(i)$ is Averaged Movement index across all heads and layers after the $i_{th}$ curriculum training phase. $h$ and $i$ are iterators $h$ and $n$ are iterators over the total number of heads $H$ and total number of layers $N$ in the BERT Transformer. Total number of curriculum phases are $c$. Since, $M$ is a difference between subsequent phases, there are a total of $c-1$ Average Movement index visualizations of size $TxT$. 

Other experimental details such as dataset, curriculum strategy, baby steps, model are all same as of the experiments explained in Section \ref{sec:5_difficult_tasks}. To recapitulate, we use SST-5 dataset with Auxiliary Model Technique in Baby Steps training procedure. The $c$ in Baby Steps training is 5 in our experiments.

\subsection{Results and Discussions}
\begin{figure}
\centering
  \includegraphics[height=9cm]{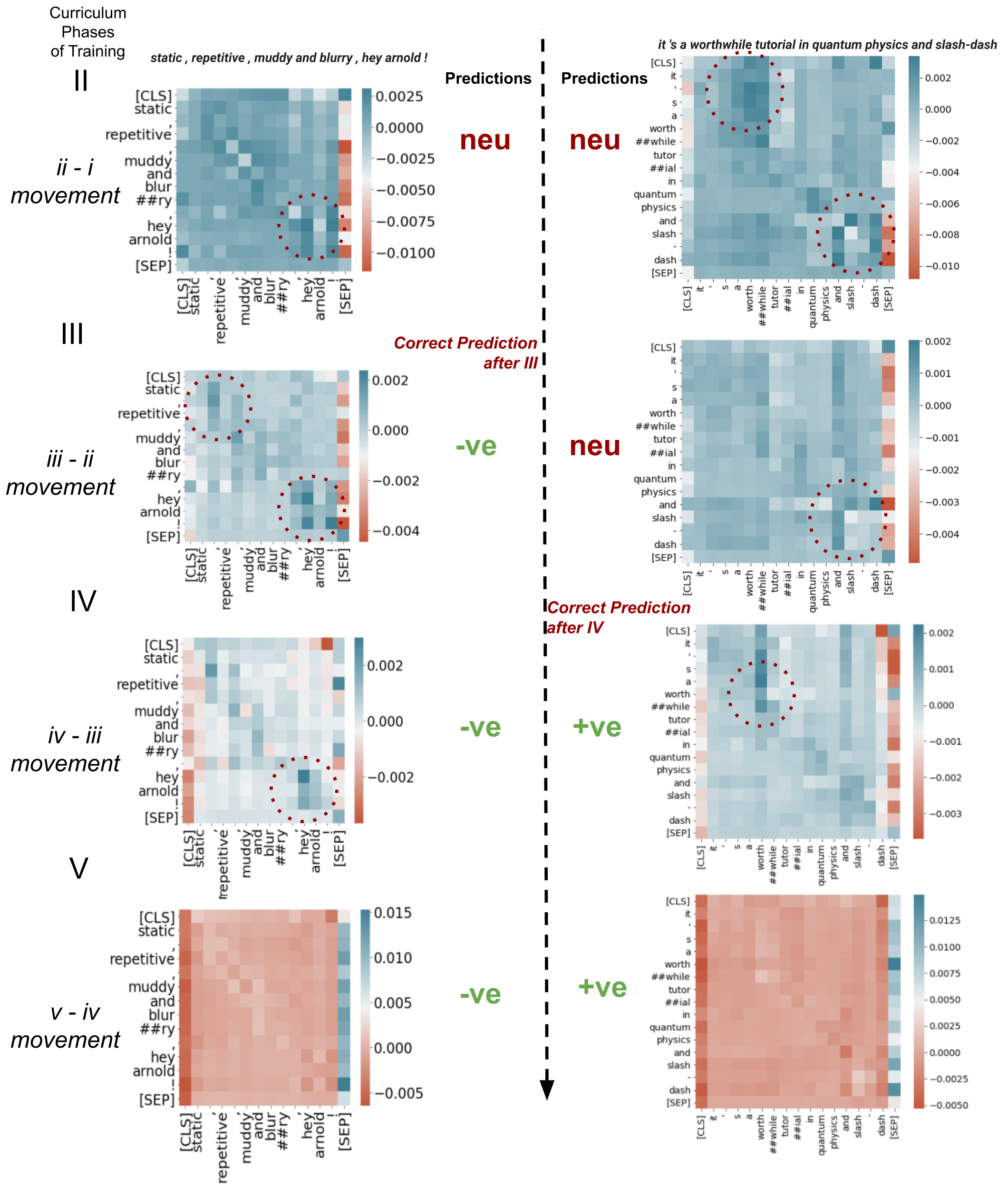}
  \caption{Attention Movement Index ($M_{avg}(i)$) visualization for the first and second sentence. The matrices show how model's focus changes for the input sentence between the phases. Blue color implies attention was added during phase transition while red implies attention diminished for those words.}
  \label{fig:viz1}
\end{figure}

Figures \ref{fig:viz1} gives examples for $M_{avg}$ scores for three sentences. There are a total five phases in our training and hence total four $M_{avg}$ scores per sentence with each score being $TxT$ in attention size with $T$ as maximum sentence length. In these images, a blue color implies there was a addition of attention or focus in the region and red or a negative value implies there was a negation in the model's focus on the area. Almost no color or white color indicates no change in attention in the region. The movement index visualizations is shown along the direction of the arrow. All the following examples are from the Testing set, after the curriculum training, the intermediate phase wise models were extracted and made to predict on the individual test sentences. Furthermore, please note that for each of the $TxT$ attentions, y-axis denotes the input positions and x-axis denotes the output positions of the self attention. Our analysis can be listed for each sentence as follows: 
 
\begin{itemize}
    \item In this sentence (True Label: Negative), \textit{``static , repetitive , muddy and blurry , hey arnold !''} the model predicts incorrectly in the beginning and continues to mislabel until  the third phase of training. We observe that on finishing the third phase of the training the movement index (2nd image from top), shows a motion of higher focus on ``static repetitive muddy'' (and lower focus from redundant words. While in the first and second phase when the model observed just a sample of easy data, the movement of attention wasn't in any direction that could help the classification. The motion in third phase however shows the addition of focus on words which could help the model classify the sentence as negative. Furthermore, until the last phase we also see an addition of focus to the terms ``hey arnold !'', this could model's way of focusing on some neutral words to avoid predicting Very Negative as opposed to Negative. Essentially, CL has divided up the task of predicting sentiment into multiple sections where the model first learnt the neutrality from ``hey arnold'' followed by focusing the polarity terms to predict just negative. This makes the prediction easier and hence effective than no curriculum.
    \item In this sentence (True Label: Positive), \textit{``it 's a worthwhile tutorial in quantum physics and slash-dash''} the model predicts incorrectly until model's completion of Phase four training. In phases two and three we see that model is adding weight to some neutral words such as ``it ' s a'' and ``slash-dash'', hence ends up predicting neutral which is an incorrect prediction. However, After observing phase four samples, model finally increases attention to the actual polarity indicating words in the sentence ``worthwhile''. and hence shifts prediction appropriately to Positive. There is a possibility that since ``worthwhile'' does not exist in BERT's vocabulary it might have found it hard to map the word to positive polarity until enough training data was observed by the model in subsequent phases. 
\end{itemize}
In all the above examples, there is a common pattern, which is that CL spreads out the sentiment prediction process. Training on initial easier samples rules out whether prediction is correct.

\section{Conclusion}
In this paper we conduct experiments to analyze CL. We first hypothesize the failure of One Pass for Curriculum learning in text classification. We conduct an Individual non curriculum experiment, to show that One Pass heavily suffers with catastrophic forgetting. Unless this forgetting is tackled, One Pass will continue to under perform even no curriculum settings. Then, we analyze CL along Task Difficulty Hypothesis to establish firmly that CL only helps when coupled with difficult tasks. The scope of improvement may diminish or even be negative for tasks which are already easy. Finally we propose movement visualizations to analyse CL. We observe that for hard examples, CL breaks down the task into multiple phases. These phases in a way rule out certain sections of the sentence for sentiment prediction first, such that when the model encounters harder data in later phases, it is clear on what not to focus on. Furthermore, after both the experiments, we can conclude that the reason CL might be deleterious to performance in easier tasks could be because in these tasks, multiple phases are unnecessary. If the model prediction was already correct in the first couple of phases, then further phases may only move the prediction away hence leading to poor performance.

\bibliography{anthology,eacl2021}
\bibliographystyle{acl_natbib}

\appendix
\section{Difficult and Easy samples according to the Auxiliary Strategy}
\label{sec:examples}
Table \ref{tab:sst5_examples} shows some examples of difficult ad easy samples according to the Auxiliary Model Curriculum Strategy explained in section \ref{sec:aux_strategy}. As we can observe, this curriculum is effective because it considers a sample as difficult in similar ways to a human discerning the sentiment as well.
\begin{table*}[h]
\begin{tabularx}{\textwidth}{l X}
    \hline
    Label &  Example\\
    \hline
     Easy  & 1. meeting , even exceeding expectations , it 's the best sequel since the empire strikes back ... a majestic achievement , an epic of astonishing grandeur and surprising emotional depth . \\
    
     (from first 50 samples) & 2. the most wondrous love story in years , it is a great film .  \\
    
     & 3. one of the best looking and stylish animated movies in quite a while ...  \\
    \hline
    Hard  & 1. if the predictability of bland comfort food appeals to you , then the film is a pleasant enough dish .   \\

    (from last 50 samples)   & 2. it is a testament of quiet endurance , of common concern , of reconciled survival .'  \\

       & 3. this movie is so bad , that it 's almost worth seeing because it 's so bad .'  \\
     \hline
     \hline
\end{tabularx}
  \caption{Examples of Difficult and Easy samples according to the Auxiliary Model Strategy}
  \label{tab:sst5_examples}
\end{table*}

\section{Additional Explanation for Attention Movement Visualization}
\begin{figure*}
\centering
  \includegraphics[width=9cm]{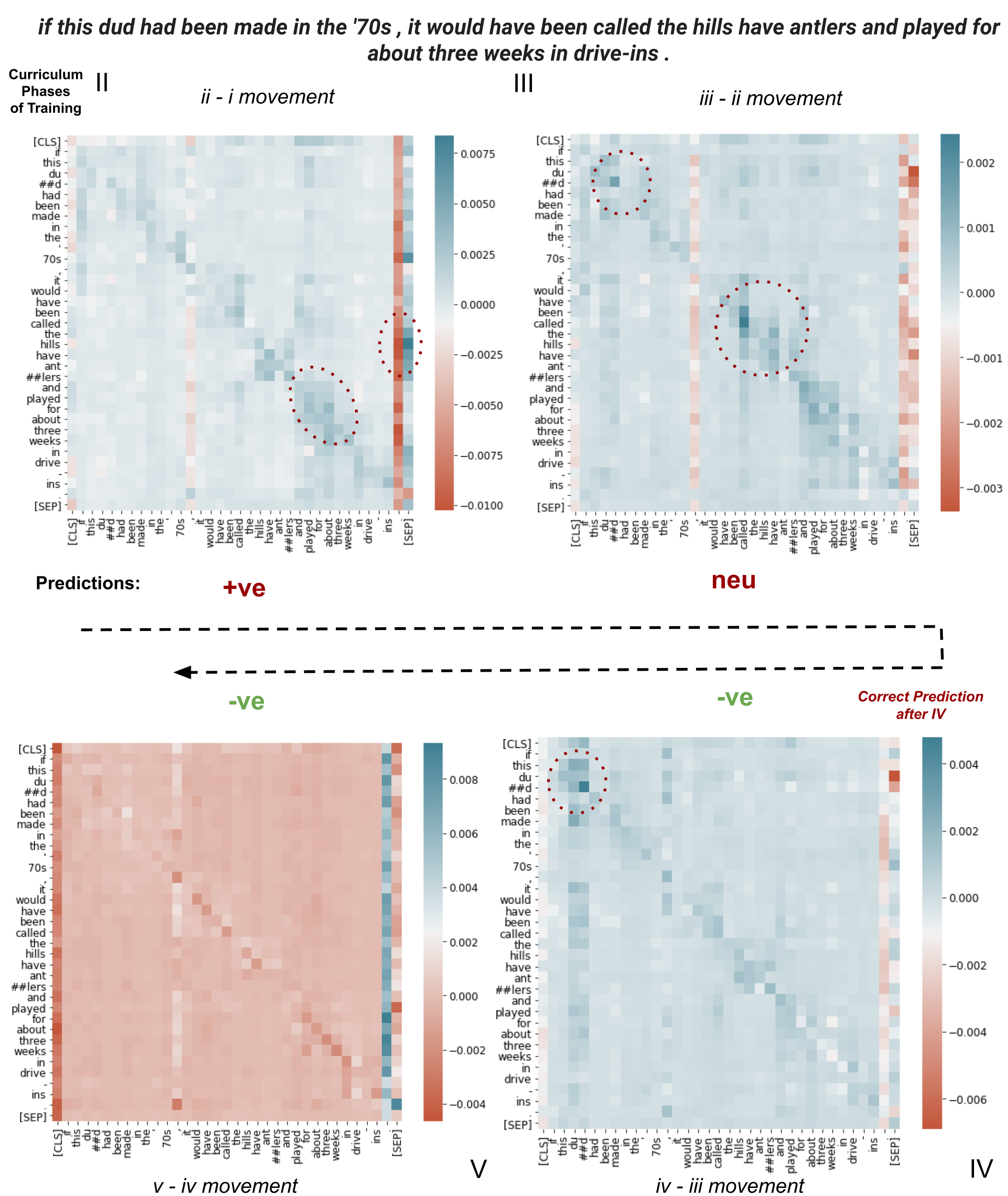}
  \caption{Attention Movement Index Visualizations for the sentence. The matrices show how model's focus changes for the input sentence between the phases. Blue color implies attention was added during phase transition while red implies attention diminished for those words.}
  \label{fig:viz2}
\end{figure*}

In the sentence from figure \ref{fig:viz2} (True Label: Negative) \textit{``if this dud had been made in the '70s , it would have been called the hills have antlers and played for about three weeks in drive-ins .''}, the model predicts incorrectly until the fourth phase. After the second phase, the model predicts Positive by increasing focus on parts of the sentence such as ``the hills'' and ``played for'' which could have a slight positive (as opposed to very positive) connotation in a movie review context. After the third phase, the motion is more towards ``the hills'' and ``dud'' which have neutral and negative connotation respectively leading to an overall sentiment of Neural. Finally, after the fourth phase, the movement is most towards the only strongly negative word ``dud''. Like the previous example, ``dud'' is not part of BERT's vocabulary and the only reason the model choose to increase its attention here must be because it already explored other regions of the sentence in previous phases. Hence, without this specific ordering, model  would have lower probability of focusing on the broken subwords ``\textit{du - \#\#d}''.\\ 
In all the examples, there is a common pattern, that curriculum learning spreads out the sentiment prediction process. Training on initial easier samples rules out incorrect predictions from parts of the sentence which do not contribute to the sentiment. This helps in later phases when model receives information to focus on right segment of the sentence. In this stage, the model would be more confident in its prediction since other regions of the sentence were already explored and excluded from the prediction process. In essence, CL debases the probability of incorrect prediction by revealing the right data at the right time in the spread out curriculum form of training.

\end{document}